\documentclass[letterpaper]{article} % DO NOT CHANGE THIS
\usepackage{aaai23}  % DO NOT CHANGE THIS
\usepackage{times}  % DO NOT CHANGE THIS
\usepackage{helvet}  % DO NOT CHANGE THIS
\usepackage{courier}  % DO NOT CHANGE THIS
\usepackage[hyphens]{url}  % DO NOT CHANGE THIS
\usepackage{graphicx} % DO NOT CHANGE THIS
\usepackage{natbib}  % DO NOT CHANGE THIS AND DO NOT ADD ANY OPTIONS TO IT
\usepackage{caption} % DO NOT CHANGE THIS AND DO NOT ADD ANY OPTIONS TO IT
\usepackage{algorithm}
\usepackage{algorithmic}
\usepackage{bm}

\usepackage{amsthm}
\usepackage{amsmath,amsfonts}

\usepackage{booktabs}
\usepackage{caption}
\usepackage{textcomp}
\usepackage{subfigure}

\usepackage{todonotes} % self
\newcommand{\hide}[1]{}

\usepackage[normalem]{ulem}

\usepackage{newfloat}
\usepackage{listings}
\DeclareCaptionStyle{ruled}{labelfont=normalfont,labelsep=colon,strut=off} % DO NOT CHANGE THIS
\floatstyle{ruled}
\newfloat{listing}{tb}{lst}{}
\floatname{listing}{Listing}
\title{Human-instructed Deep Hierarchical Generative Learning for Automated Urban Planning}

\author {
        Dongjie Wang\textsuperscript{\rm 1}
        Lingfei Wu \textsuperscript{\rm 2}
        Denghui Zhang \textsuperscript{\rm 3}
        Jingbo Zhou \textsuperscript{\rm 4}
        Leilei Sun \textsuperscript{\rm 5}
        Yanjie Fu \textsuperscript{\rm 1}\footnote{Contact Author}
       \\
}
\affiliations {
    \textsuperscript{\rm 1}  University of Central Florida 
    \textsuperscript{\rm 2} Pinterest 
    \textsuperscript{\rm 3} Rutgers University 
    \textsuperscript{\rm 4} Baidu Research
    \textsuperscript{\rm 5}  Beihang University
    \\
    wangdongjie@knights.ucf.edu, lwu@email.wm.edu, dhzhangai@gamil.com,\\ zhoujingbo@baidu.com, leileisun@buaa.edu.cn, yanjie.fu@ucf.edu
}

\begin{document}

\maketitle

\begin{abstract}
%Our intuition is that a urban planning configuration can be summarized by a few handfuls of functionality projections. To leverage this intuition, we introduce a functionalizer module to to convert embeding of human instructions and sorrounding contexts into compact sets of functionalitiy projections.

%Urban planning is to generate an optimized land-use configurations for a target area.

The essential task of urban planning is to generate the optimal land-use configuration of a target area.
However, traditional urban planning is time-consuming and labor-intensive.
Deep generative learning gives us hope that we can automate this planning process and come up with the ideal urban plans.
While remarkable achievements have been obtained, they have exhibited limitations in lacking awareness of: 1) the hierarchical dependencies between functional zones and spatial grids; 2) the peer dependencies among functional zones; and 3) human regulations to ensure the usability of generated configurations.
To address these limitations, we develop a novel human-instructed deep hierarchical generative model.
We rethink the urban planning generative task from a unique functionality perspective, where we summarize planning requirements into different functionality projections for better urban plan generation. 
To this end, we develop a three-stage generation process from a target area to zones to  grids. 
The first stage is to label the grids of a target area with latent functionalities to discover functional zones. 
The second stage is to perceive the planning requirements to form urban functionality projections.
We propose a novel module: functionalizer to project the embedding of human instructions and geospatial contexts to the zone-level plan to obtain such projections.
Each projection includes the information of land-use portfolios and the structural dependencies across spatial grids in terms of a specific urban function.
The third stage is to leverage multi-attentions to model the zone-zone peer dependencies of the functionality projections to generate grid-level land-use configurations.
Finally, we present extensive experiments to demonstrate the effectiveness of our framework.  

\end{abstract}

\section{Introduction}
% 1. human instruction order, affect results
% 2. space hierarchy, functional region - land use configuration mapping,high-level (fine-grained) -> low level (coarse-grained)
% 3. planning correlation (complementary, dependency) between regions

% explicit human involve
% affiliative relationship

% Pengyang
Urban planning is vital for building up a sustainable and vigorous community. 
As a complicated and time-consuming task, traditional practice heavily depends on experts' personal experiences. 
The variance among urban planners may result in biases and implausible solutions. 
Thanks to the explosive development of deep learning and internet-of-things, the handful of methodologies and ubiquitously available geo-social, urban and mobile data provide us with a new data-driven perspective to re-investigate urban planning.

\begin{figure}[!t]
    \centering
    \includegraphics[width=1.0\linewidth]{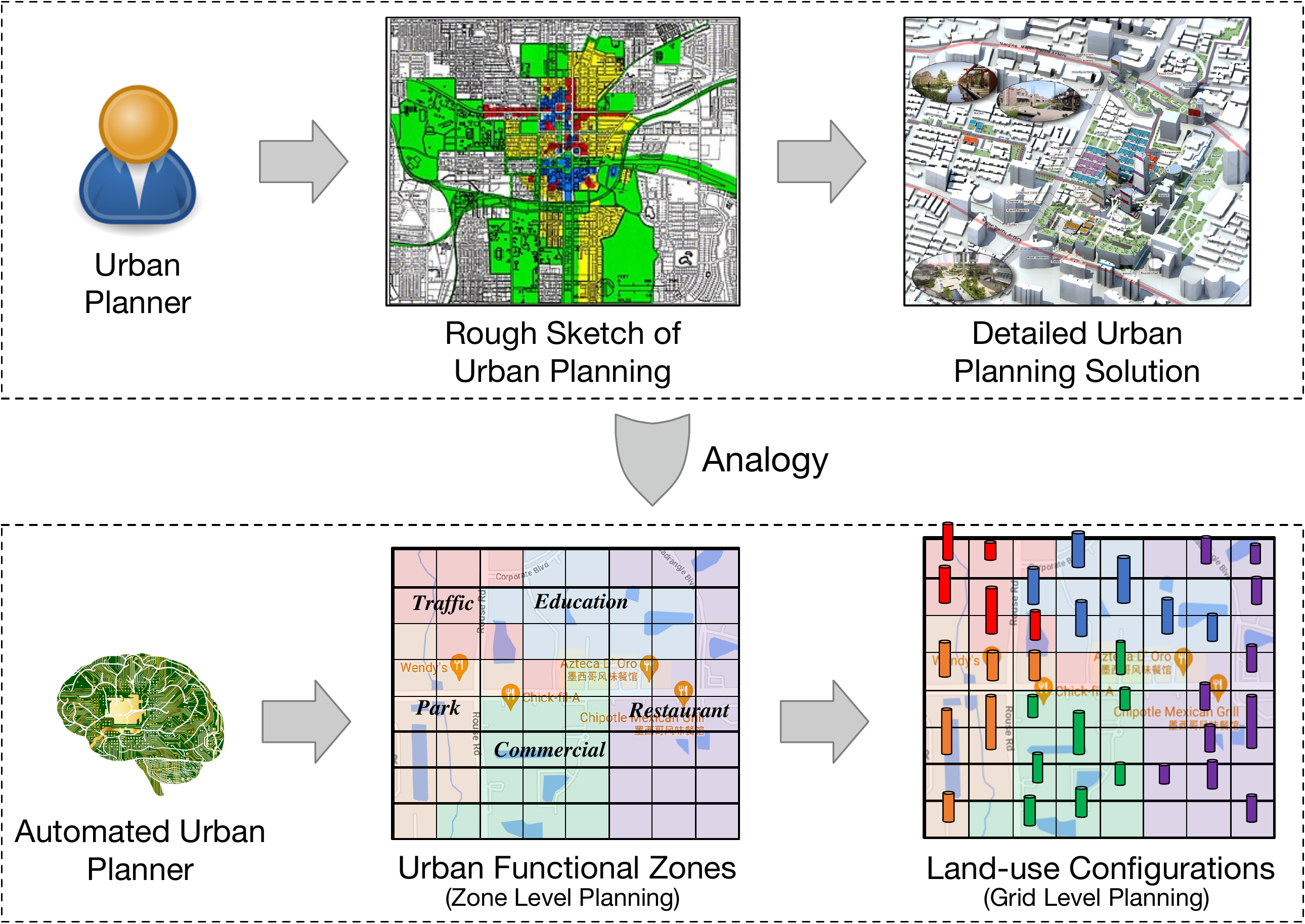}
    % \vspace{-0.2cm}
    \captionsetup{justification=centering}
    % \vspace{-0.2cm}
    \caption{
    The automated urban planner can mimic the workflow of urban experts by first generating zone-level planning and then refining it to grid-level planning.
    }
    % \vspace{-0.7cm}
    \label{fig:framework_motivation}
\end{figure}

There are considerable existing works related to automated urban planning
~\cite{wang2021automated,shen2020machine,ye2021masterplangan,wang2021deep}.
For example, motivated by the remarkable success of deep image generation, ~\cite{wang2020reimagining} proposes a land-use configuration generation framework, namely LUCGAN, which can generate a land-use configuration automatically for an empty geographical area based on surrounding contexts. 
While the existing works have achieved promising results, there are still several limitations:
1) hierarchical relationships between high-level urban functional zones and the detailed urban planning scheme are ignored;
2)mutual dependencies and influences among the planning of different subareas are omitted.
3) human instructions from planning experts, such as safety level, greening rate, volume rate, and etc,  cannot be perceived by model;

Therefore, in this paper, we study the research problem of how to employ deep models to make  automated  urban planning more intelligent. 
To settle the problem, we can formulate urban planning as a deep conditional generative task, in which human instructions and surrounding contexts can be regarded as the generative condition, and spatial hierarchical relationships and planning dependencies can be considered as the generative constraints. The objective is to generate an urban solution constrained by various factors.

However, there are three unique challenges in the defined generative task:
1) \textbf{Challenge 1: Capturing Spatial Hierarchical Relations}: urban functional zones reflect the land-use layout of a geographical area, which provides the bedrock of a land-use configuration.
Neglecting such spatial hierarchical relations between urban functional zones and the land-use configuration may result in the unstable generation performance.
But how can we model the spatial hierarchies during the generation process?
2) \textbf{Challenge 2: Capturing Planning Dependency Among Subareas}:
planning solutions of different sub geographical areas are mutually dependent and affected. 
In a geographical area, for example, if some subareas have been built up with a lot of business buildings, the other subareas will be built up with more entertainment as a supplement to the urban functions.
But how can we capture the planning dependencies among different subareas?
3) \textbf{Challenge 3: Integrating Human instructions from Planning Experts}: urban planning is a highly complicated and personalized task.
To produce plausible urban planning results,
planning experts always consider various realistic factors (\textit{e.g.} greening rate level, safe level, volume rate level).
But how can we integrate such human instructions for improving the personalized generative capability of the model?

To tackle the above challenges, we propose a novel Human-\textbf{I}nstructed Deep \textbf{H}ierarchical Generative Framework (\textbf{IHPlanner}), which can generate a desired land-use configuration for an empty area based on human instructions and surrounding contexts, as well as considering the spatial hierarchies and the planning dependencies.
Our main contributions can be summarized as follows:
1) \textbf{Formulating the automated urban planning as a multi-scale generation framework}.
The classical workflow of urban experts is to first design a rough sketch, then fill concrete designing elements to obtain the final urban plan.
Imitating such a designing workflow, the proposed multi-scale generation framework generates the coarse-grained skeleton (urban functional zones)  at the first stage, and then produces the fine-grained urban plan (land-use configuration) based on the skeleton at the second stage.
This framework setting automatically captures the spatial hierarchies between urban functional zones and land-use configurations.
2) \textbf{Involving human instructions from planning experts via conditional embedding}. 
We formulate the human instructions from experts as the generative condition in urban plan generation.
% The urban planning of an empty geographical area is not only determined by the subjective needs of urban experts, but also affected by socioeconomic characteristics (\textit{e.g.} transportation, demography, economy) of surrounding contexts.
To make our model perceive these conditions, we convert them into embedding vectors.
To control the generation process,  we concatenate such embedding vectors and regard them as the model input.
3) \textbf{Semantic segmentation-based generation to capture planning dependency}.
% The POI distribution of different subareas are affected by each other. 
Human instructions and surrounding contexts contain enormous semantics that implicitly reflect the planning requirements for coarse-grained urban functional zones.
Therefore, we design a planning semantic segmentation module to allocate the corresponding semantics to each urban functional zone respectively. 
We exploit the multi-head attention mechanism ~\cite{vaswani2017attention} to capture the dependencies among segmented semantics for quantifying the planning dependencies among subareas.
% To capture the planning dependency, we dispatch the semantics of planning requirements of human instruction and surrounding contexts into different subareas to form the semantically-rich urban function tokens.
% After that, we utilize the multi-head attention mechanism ~\cite{vaswani2017attention} to capture the semantic dependencies among these tokens.
Moreover, self-designed planning layers are developed to generate the final land-use configuration.
4) \textbf{Extensive experiments and case studies to validate the effectiveness of our framework}.
We conduct all experiments and case studies based on the geographical data, traffic flow, road map, POIs, and check-in records of Beijing. 
We compare our proposed framework with six state-of-the-art deep generative models, and provide visualization to show the superiority of our framework.

\section{Preliminaries}
% We will introduce necessary definitions and the problem statement.

% \vspace{-0.1cm}
\subsection{Definitions}

\begin{figure}[!t]
    % \vspace{-0.4cm}
    \subfigure[Geospatial contexts encircle the target area from different directions.  ]{\includegraphics[width=3.6cm]{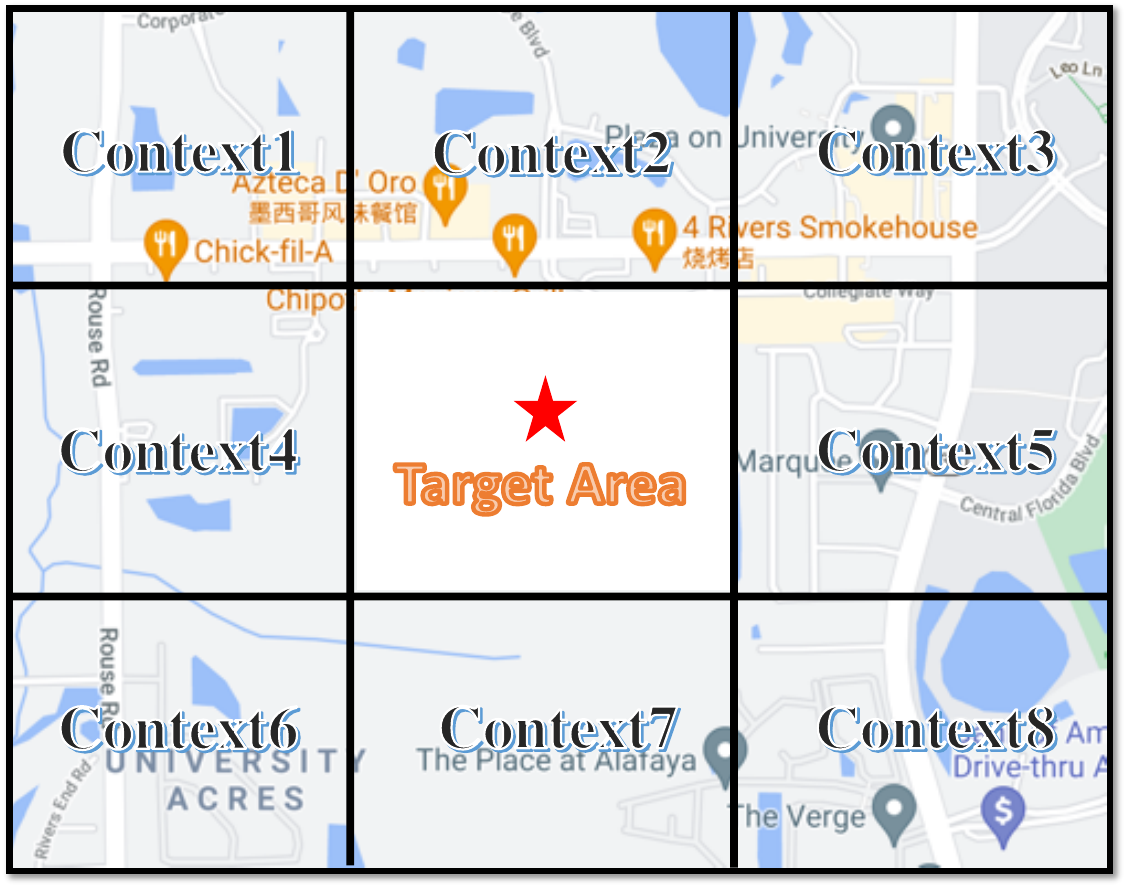} \label{fig:target_area}}\hspace{3mm}
    \subfigure[The spatial attributed graph contains all features of geospatial contexts. ]{\includegraphics[width=3.8cm]{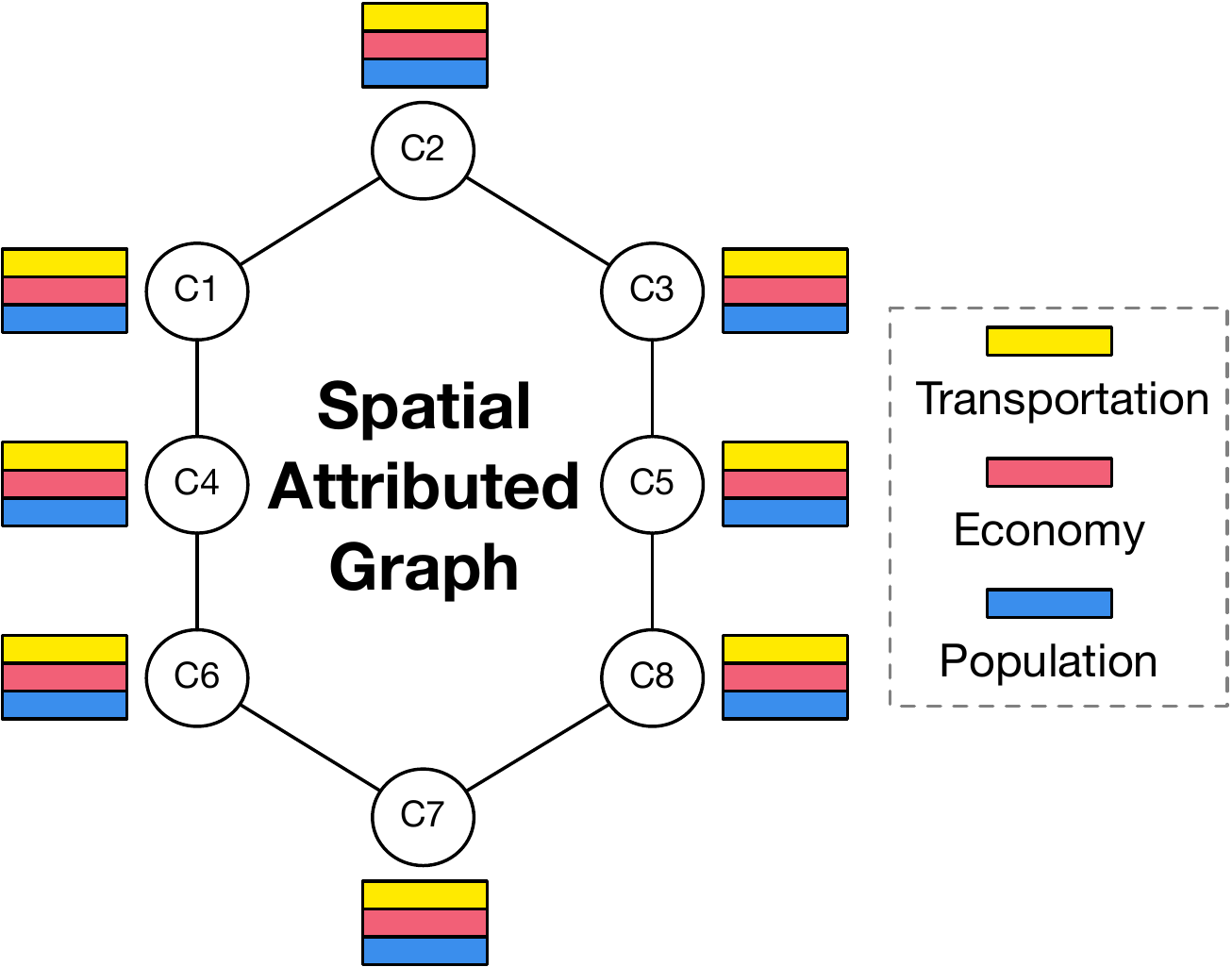}\label{fig:sag}}
    % \vspace{-0.35cm}
    \caption{Illustration of target area and geospatial contexts.}
    \label{fig:target_geo}
    % \vspace{-0.2cm}
    \end{figure}

    \begin{figure}[!t]
    % \vspace{-0.15cm}
    % \subfigure[Zone level planning provides a planning foundation for grid level planning.]
    \subfigure[Zone-level planning is a 2-D matrix, which provides a high-level guidance for grid-level planning.]
    {\includegraphics[width=3.9cm]{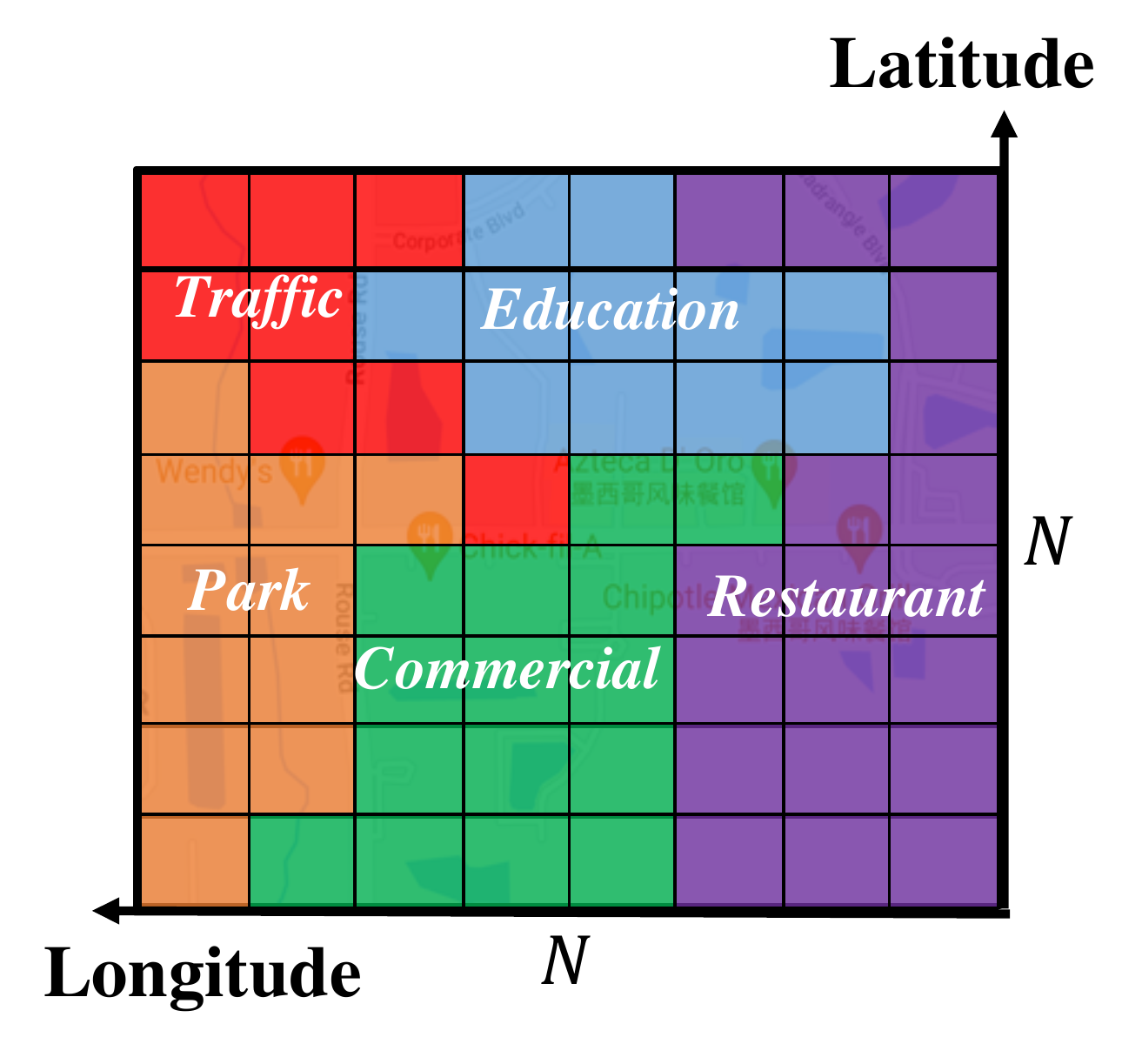} \label{fig:urban_func}}\hspace{3mm}
    % \subfigure[Grid level planning is a 3 dimensional tensor (longitude, latitude, POI category).]
    \subfigure[Grid-level planning is represented by a 3-D tensor where we reserve the 3rd dimension for POI as each grid may contain multiple POI categories.]
    {\includegraphics[width=4.0cm]{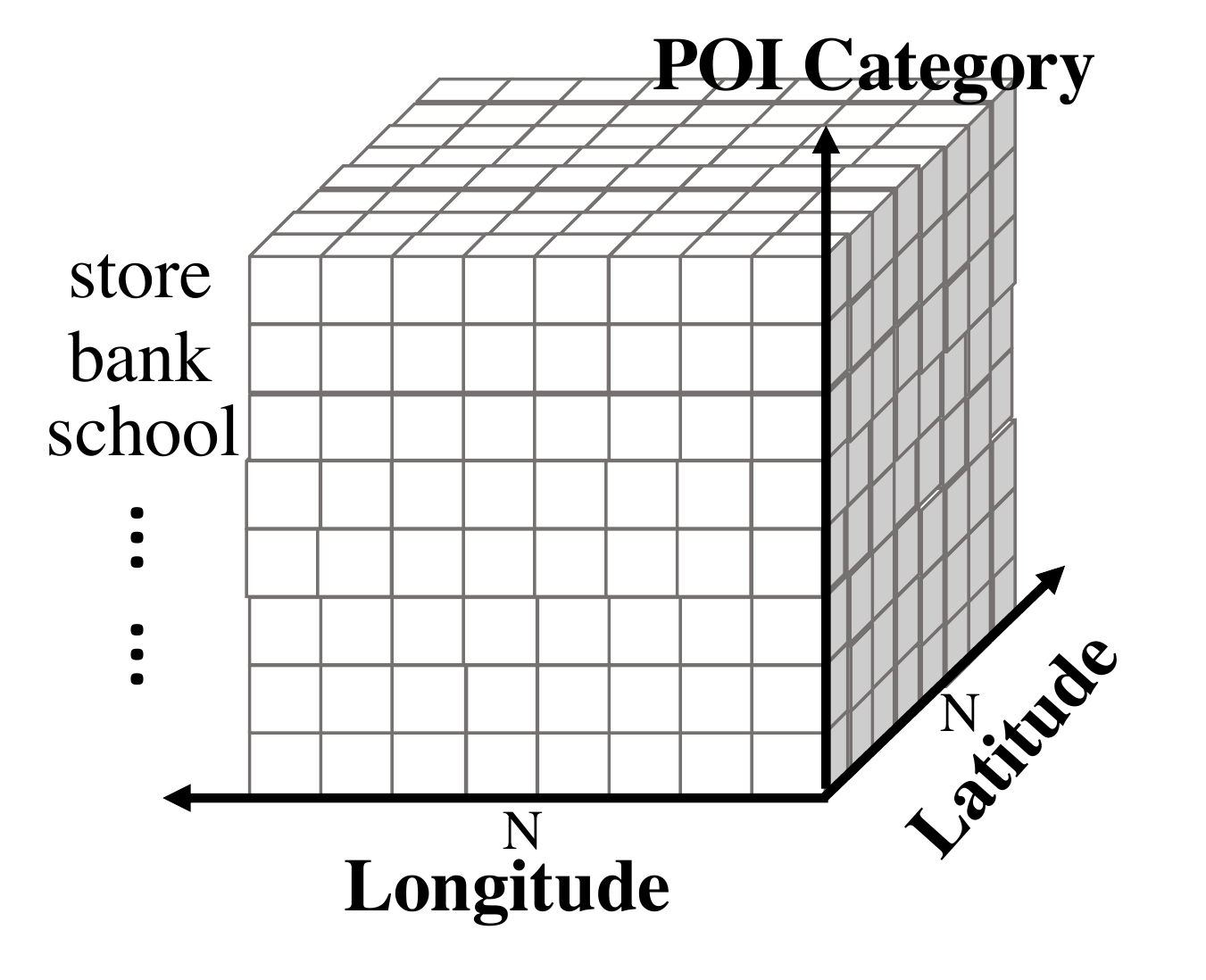}\label{fig:land-use}}
    % \vspace{-0.25cm}
    \caption{Urban functional zone and land-use configuration.}
    \label{fig:urban_land}
    % \vspace{-0.6cm}
\end{figure}

\begin{figure*}[!t]
% \vspace{-0.15cm}
    \centering
    \includegraphics[width=1.0\linewidth]{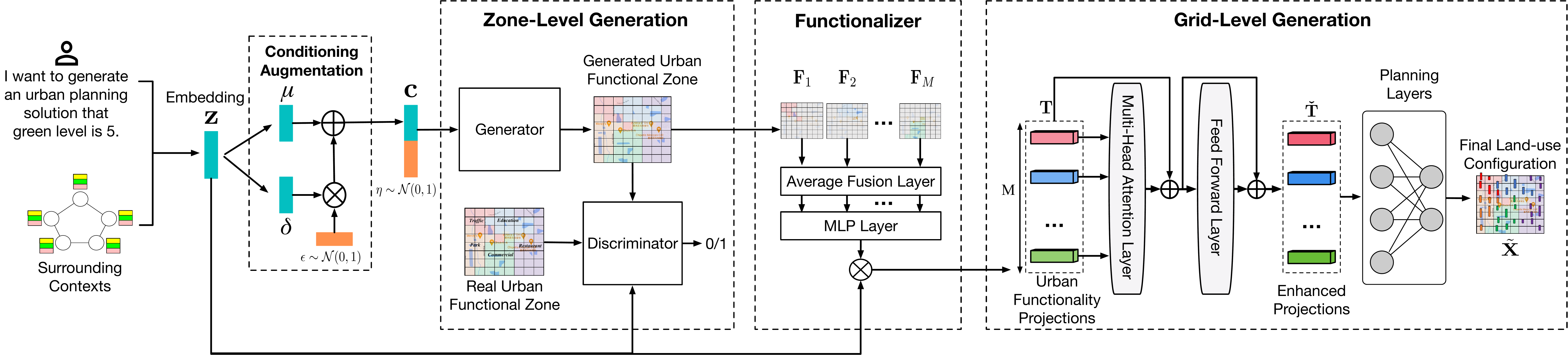}
    % \vspace{-0.15cm}
    \captionsetup{justification=centering}
    % \vspace{-0.4cm}
    \caption{
    The overview of IHPlanner.
    It consists of four main steps: Conditioning Augmentation, Zone-level Generation,
    Functionalizer,
     and Grid-level Generation.
    }
    % \vspace{-0.5cm}
    \label{fig:framework}
\end{figure*}

\textbf{Target Area and Geospatial Contexts.}
Target area is an empty and square geographical region (\textit{e.g.} a square with a side length of 1 kilometer). 
Geospatial contexts are the surrounding environments, each of which has the same shape as the target area.
Figure \ref{fig:target_area} illustrates that geospatial contexts encircle the target area from different directions.
To leverage the information of geospatial contexts ~\cite{wang2020reimagining}, we formulate such contexts as a spatial attributed graph described in Figure ~\ref{fig:sag}.
In this graph,
a node is a geographical region, an edge reflects the spatial connectivity between any two regions, and the attributes of a node are the socioeconomic features of the corresponding region.

    % \begin{figure}[!t]
    % \centering
    % \includegraphics[width=0.8\linewidth]{img/urban_func.pdf}
    % \vspace{-0.4cm}
    % \captionsetup{justification=centering}
    % % \vspace{-0.13cm}
    % \caption{Zone level planning provides a rough sketch of the urban planning of a geographical area.}
    % \vspace{-0.4cm}
    % \label{fig:urban_func}
    % \end{figure}
    
    % \begin{figure}[!t]
    % \centering
    % \includegraphics[width=0.9\linewidth]{img/land_use.pdf}
    % \vspace{-0.4cm}
    % \captionsetup{justification=centering}
    % \vspace{-0.1cm}
    % \caption{The collection process of the grid level planning.}
    % \vspace{-0.55cm}
    % \label{fig:land_use}
    % \end{figure}

\textbf{Urban Functional Zone and Land-use Configuration.} 
\label{ufz_a_luc}
In this paper, urban functional zones (\textit{i.e.} zone-level planning) provide a planning foundation for the land-use configuration (\textit{i.e.} grid-level planning).
\textbf{For the urban functional zone}, following the idea in studies ~\cite{yuan2014discovering}, we utilize the geographical data and human mobility to extract.
Specifically,  we first divide a geographical area into $N \times N$ grids. Then, we consider the grids to be words and the human trajectories to be sentences, and so all trajectories inside the area constitute a document.
Next, we use a topic model to discover the specific urban function label of each grid to obtain the final results.
Figure ~\ref{fig:urban_func} shows the data structure of such zones.
The zone-level planning is a matrix, denoted by $\mathbf{U}\in \mathbb{R}^{N \times N}$, in which multiple grids affiliate to one urban function label.
\textbf{For the land-use configuration}, we adopt the quantitative definition in studies ~\cite{wang2020reimagining}.
Specifically, we first split a geographical area into $N\times N$ grids. 
Then, we count how many POI (\textit{i.e.} Point of Interest) locate in each grid under different POI categories.
After that, we stack these counted results together as the final configuration.
Figure ~\ref{fig:land-use} shows the data structure of such configuration, which is a tensor consisting of longitude, latitude, and POI category dimensions.
The tensor is
denoted by $\mathbf{\widehat{X}} \in \mathbb{R}^{N \times N \times C}$, where $C$ is the number of POI categories.

\textbf{Human Instruction.}
In this paper, human instruction is to guide the generation process of our planning framework. To allow our model to perceive such instruction,
we quantify its semantic meaning into different levels.
For instance, the range of green rate (\textit{i.e.} the coverage of green plants of a geographical area) is $[0\sim 1]$.
We divide the green rate into multiple green rate levels.
The label of these green rate levels is human instruction.

% Human experts can input the label of green rate level into our planner to indicate their planning requirements.
% So, these labels of green rate level refer to human instructions.

\subsection{Problem Statement}
Our goal is to develop an automated urban planner, which can generate a land-use configuration for an empty target area based on human instructions and geospatial contexts.
Formally, given geospatial contexts denoted by $\mathcal{G}$, human instructions denoted by $I$,  land-use configurations denoted by $\mathbf{\widehat{X}}$,
we aim to find a mapping function $f:(\mathcal{G}, I) \rightarrow \mathbf{\widehat{X}}$.
The function $f$
takes geospatial contexts $\mathcal{G}$ and human instructions $I$ as input, and outputs the corresponding grid-level land-use configuration $\mathbf{\widehat{X}}$.

% \vspace{-0.1cm}
\section{Methodology}

% We present an overview and  technical components of our  method. 
% , as well as comparison with prior literature

% \vspace{-0.2cm}

\subsection{Framework Overview}
Figure \ref{fig:framework} shows the  overview of our framework IHPlanner.
The pipeline framework has four key components: \textbf{conditioning augmentation, zone-level generation, functionalizer, and grid-level generation}.
Specifically, for an empty target area, we first preserve the planning requirements contained in human instructions and geospatial contexts into an embedding vector.
Then, considering the data sparsity issue, we utilize the conditioning augmentation module to  increase the data diversity.
Next, we employ the zone-level generation module to generate the zone-level planning that provides a planning foundation for the grid-level generation.
After that, in the functionalizer module,  we project the semantics of planning requirements into different functional zones to obtain the urban functionality projections.
This projection process converts the planning dependencies across functional zones into semantic correlations among these projections.
Finally, in the grid-level generation module, we use multi-attentions to capture such semantic correlations, then employ planning layers to generate the grid-level planning.

% \vspace{-0.1cm}
\subsection{Conditioning Augmentation}
The dataset for automated urban planning is sparse,
resulting in model overfitting or terrible generation performance of IHPlanner.
We adopt the conditioning augmentation module to mitigate the learning issue.
To make our method comprehend the planning semantics included in human instructions and surrounding geospatial contexts, we first convert the spatial attributed graph extracted from geospatial contexts into a graph embedding by ~\cite{kipf2016variational, wu2021graph,wu2022graph}, and then concatenate it with the one-hot vector of human instructions as the model input.

To be convenient, we adopt the $k$-th empty target area to explain the following calculation process.
Specifically, we denote $\mathbf{z}^{(k)} \in \mathbb{R}^ {1 \times O}$ as the concatenated  embedding of human instructions and geospatial contexts, where $O$ is the size of the feature dimension.
We first utilize the conditioning augmentation module to estimate the distribution of $\mathbf{z}^{(k)}$. 
Then, we randomly sample an augmented embedding $\mathbf{c}^{(k)}$ from the distribution, and regard it as the input of the zone-level generation module.
The prior format of the estimated distribution is a normal distribution, denoted by $\mathcal{N} (\mu(\mathbf{z}^{(k)}),\delta(\mathbf{z}^{(k)}))$, where $\mu(.)$ and $\delta(.)$ indicate the mean and covariance function respectively.
The mean and covariance value of the distribution are updated over learning process.
We adopt the reparameterization technique to imitate the sampling operation, which can be formulated as follows:
 \begin{equation}
%  \vspace{-0.1cm}
    \mathbf{c}^{(k)} =\bm{\mu}(\mathbf{z}^{(k)})+\bm{\delta}(\mathbf{z}^{(k)}) \times \epsilon
\end{equation}
where $\epsilon$ is a random variable vector sampled from a standard normal distribution $\mathcal{N}(0,1)$.

% \vspace{-0.1cm}
\subsection{Zone-level Generation}
\label{cg_part}
Inspired by the workflow of human planners, we can first generate a rough sketch of urban planning, then refine the sketch to the grid-level land-use configuration.
Specifically, for the $k$-th empty target area, we first concatenate the embedding $\mathbf{c}^{(k)}$ and the random variable embedding $\bm{\eta}^{(k)}$ together, then input it into a generator to generate the urban functional zones.
Here, $\bm{\eta}^{(k)}$ is sampled from the standard normal distribution $\mathcal{N}(0,1)$, which can improve the robustness and generalization of model.
Next, we combine the generated result and the embedding $\mathbf{z}^{(k)}$ together, then input it into a discriminator.
The discriminator is to justify whether the input is the combination of the real urban functional zones $\mathbf{U}^{(k)}$ and $\mathbf{z}^{(k)}$.
We alternatively optimize the generator and the discriminator until model convergence.

When optimizing the generator, we minimize equation \ref{loss_g}:
\begin{equation}
\begin{split}
    \mathcal{L}_G = \sum_{k=1}^K log(1-D(G(\bm{\eta}^{(k)},\mathbf{c}^{(k)}),\mathbf{z}^{(k)}))
    \\
    + 
    \lambda \cdot KL[\mathcal{N} (\mu(\mathbf{z}^{(k)}),\delta(\mathbf{z}^{(k)})) || \mathcal{N}(0,1)],
\end{split}
\label{loss_g}
\end{equation}
where $KL[.]$ indicates the Kullback-Leibler (KL) divergence between the distribution $\mathcal{N} (\mu(\mathbf{z}^{(k)}),\delta(\mathbf{z}^{(k)}))$ and a standard normal distribution  $\mathcal{N}(0,1)$; 
$\lambda$ is a scalar, which adjusts the contribution of the item $KL[.]$ in $\mathcal{L}_G$.
$\mathcal{L}_G$ can be divided into two parts by  "$+$".
Intuitively,
the first part tries to minimize the differences between the generated zone-level planning and the real zone-level planning, which improves the generation performance of the generator gradually.
The second part tries to smooth the distribution $\mathcal{N} (\mu(\mathbf{z}^{(k)}),\delta(\mathbf{z}^{(k)}))$ produced by the conditioning augmentation module, which improves the diversity and quality of the input embedding $\mathbf{c}^{k}$  ~\cite{larsen2016autoencoding}.

When optimizing the discriminator, we maximize equation \ref{loss_d}:
\begin{equation}
\begin{split}
    \mathcal{L}_D = \sum_{k=1}^K 
    log(1-D(G(\bm{\eta}^{(k)},\mathbf{c}^{(k)}),\mathbf{z}^{(k)}))\\
    +
    log D(\mathbf{U}^{(k)},\mathbf{z}^{(k)}).
\end{split}
\label{loss_d}
\end{equation}
Intuitively, $\mathcal{L}_D$ improves the discrimination ability by urging the discriminator to provide lower scores for the generated results and evaluate higher scores for the real standards.

Ultimately, the well-trained generator can generate a suitable rough sketch of urban planning $ \mathbf{\Check{U}}^{(k)} \in\mathbb{R}^{N\times N}$ for the $k$-th empty area according to human instructions and geospatial contexts.
Each value in $\mathbf{\Check{U}}^{(k)}$ indicates the urban functionality label of the associated geographical location.

\begin{figure}[!t]
%   \vspace{-0.2cm}
    \centering
    \includegraphics[width=0.95\linewidth]{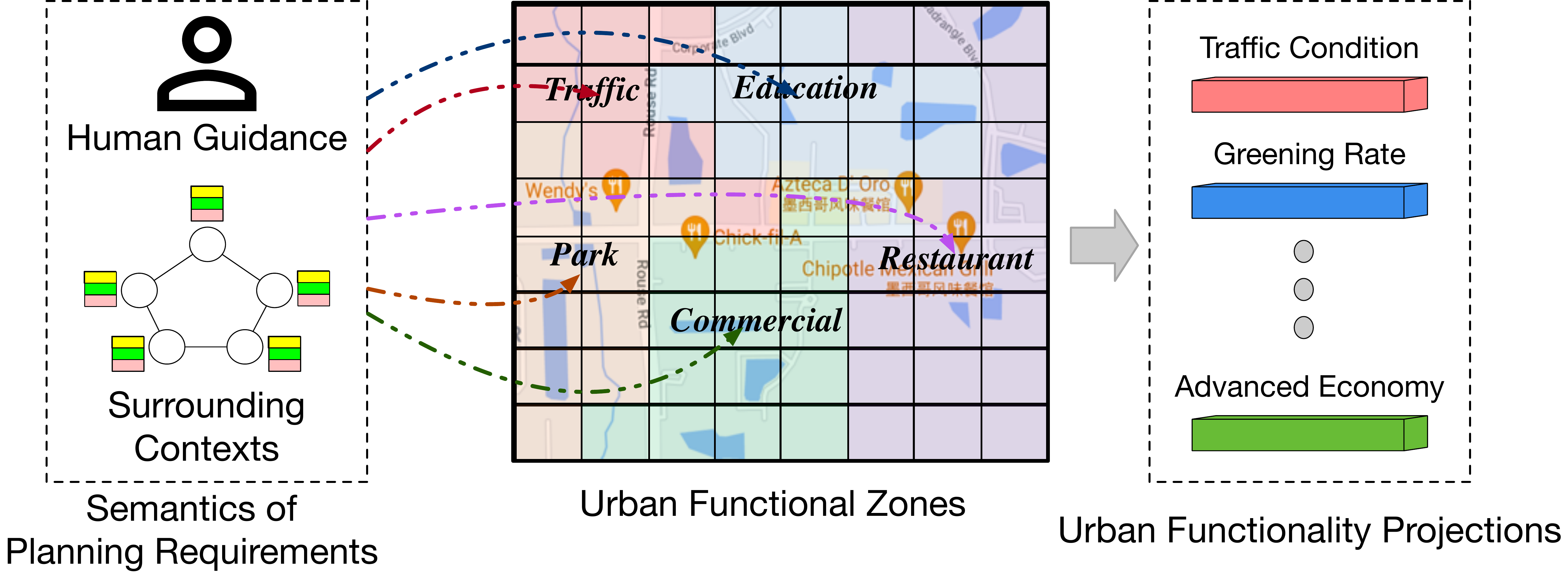}
    % \vspace{-0.1cm}
    \captionsetup{justification=centering}
    % \vspace{-0.1cm}
    \caption{The information of planning requirements is projected into different urban functional zones to form urban functionality projections.}
    % \vspace{-0.6cm}
    \label{fig:plan_segment}
    \end{figure}

% \vspace{-0.1cm}
\subsection{Functionalizer}
An outstanding urban plan can be summarized by a few handfuls of
urban functionalities such as convenient transportation, high green rate, and developed economy.
In other words, to produce such an urban plan, our planning model should consider the planning requirements on these urban functionality sides.
Thus, as illustrated in Figure ~\ref{fig:plan_segment}, we project the planning requirements contained in human instructions and geospatial contexts into different functional zones to obtain urban functionality projections.
This projection process lays the cornerstone for capturing the planning dependencies across different functional zones.

Specifically, for the $k$-th empty target area, we have generated the zone-level planning $\mathbf{\Check{U}}^{(k)}$.
Then, we divide the area into $M$ zones according to the urban function labels in $\mathbf{\Check{U}}^{(k)}$, denoted by $\mathbf{F}^{(k)}=[\mathbf{F}^{(k)}_1, \mathbf{F}^{(k)}_2, \cdots, \mathbf{F}^{(k)}_M]$, and $\mathbf{F}^{(k)}\in \mathbb{R}^{M\times N \times N}$.
Next, we calculate the semantic proportion of planning requirements for each functional zone.
After that, we multiply $\mathbf{z}^{(k)}$ with these semantic proportions to obtain urban functionality projections.
The projection process can be formulated as follows:
\begin{equation}
    \mathbf{T}^{(k)} = \text{Softmax}(\text{AVG\_Fusion}(\mathbf{F}^{(k)})\cdot \mathbf{W}_a)\cdot \mathbf{z}^{(k)},
\end{equation}
where $\text{AVG\_Fusion}(.)$ column-wisely averages the information of each functional zone respectively, which changes the shape of $\mathbf{F}^{(k)}$ to $\mathbb{R}^{M \times N}$; $\mathbf{W}_a \in \mathbb{R}^{N \times 1}$ is the weight matrix; $\text{Softmax}(.)$ outputs the semantic proportion value;
$\mathbf{T}^{(k)} \in \mathbb{R}^{M \times O}$ are the final urban functionality projections, which implicitly reflect the planning requirements under different urban functionalities.

% \vspace{-0.2cm}
\subsection{Grid-level Generation}
\label{fg_part}
Urban infrastructures and buildings in different functional zones are mutually dependent.
For instance, if several functional zones have been planned with many commercialized buildings, planners will not put the same buildings in the nearby zones but instead add entertainment facilities to increase urban vibrancy.
To capture such dependencies, we apply the multi-attentions \cite{vaswani2017attention} on urban functionality projections to obtain enhanced projections.
Then, we input these enhanced projections into  planning layers to produce the land-use configuration.

Specifically, for the $k$-th empty area, we input the urban functionality projections $\mathbf{T}^{(k)}$ into a multi-head attention layer to calculate the attention weight matrix.
The multi-head attention layer consists of $h$ single scaled dot-product attention layers.
For a single attention layer, the calculation process is as follows:
\begin{equation}
    \mathbf{A} = \text{Softmax}(\frac{\mathbf{Q}\cdot \mathbf{K}^T)}{\sqrt{d_k}})\cdot \mathbf{V},
\end{equation}
where $\mathbf{A}\in \mathbb{R}^{M\times O}$ is the attention matrix; $\mathbf{Q}, \mathbf{K},\mathbf{V}$ are the query, key, and value matrix respectively. The three matrices all come from $\mathbf{T}^{(k)}$; 
$d_k$ is the scaling factor;
$\mathbf{Q}\cdot \mathbf{K}^T \in \mathbb{R}^{M\times M}$, which indicates the semantic similarity between any two of urban functionalities;
These $h$ single attention layers have different $\mathbf{Q}, \mathbf{K},\mathbf{V}$ matrices.
These layers extract features from different  semantic representation subspaces.
Then we collect the attention weights of $h$ layers together and add  $\mathbf{T}^{(k)}$ to obtain $\mathbf{T'}^{(k)} \in \mathbb{R}^{M\times O}$,
\begin{equation}
    \mathbf{T}'^{(k)} = \mathbf{T}^{(k)} + \text{Concat}(\mathbf{A}^{(k)}_1,\mathbf{A}^{(k)}_2,\cdots,\mathbf{A}^{(k)}_h) \cdot \mathbf{W}_T,
\end{equation}
where $\mathbf{W}_T \in \mathbb{R}^{hO\times O}$ is the projection weight matrix.
After that, we utilize a fully connected feed-forward network constituted by two linear layers to attain the enhanced projections $\mathbf{\Check{T}}^{(k)} \in \mathbb{R}^{M \times O}$,
\begin{equation}
    \mathbf{\Check{T}}^{(k)} = \mathbf{T}'^{(k)} + \text{Relu}(\mathbf{T}'^{(k)} \cdot \mathbf{W}_1)\cdot \mathbf{W}_2, 
\end{equation}
where $\mathbf{W}_1, \mathbf{W}_2 \in \mathbb{R}^{O\times O}$ are two weight matrices and $\text{Relu}$ denotes the nonlinear transformation function.
Next, we input $\mathbf{\Check{T}}^{(k)}$ into planning layers to generate the final land-use configuration $\mathbf{\tilde{X}}^{(k)} \in \mathbb{R}^{N\times N\times C}$.
This process can be formulated as,
\begin{equation}
    \mathbf{\tilde{X}}^{(k)} = \mathbf{W}_u \cdot \mathbf{\Check{T}}^{(k)} \cdot \mathbf{W}_d + \mathbf{b},
\end{equation}
where $\mathbf{W}_u \in \mathbf{R}^{N \times M}, \mathbf{W}_d \in \mathbb{R}^{O \times (N\times C)}$ are the weight matrices.
During the generation process,
$\mathbf{W}_u$ aims to consider the correlations among different enhanced projections;
$\mathbf{W}_d$ aims to exploit the dependencies among different latent dimensions in these enhanced projections.
$\mathbf{b}\in \mathbb{R}^{N\times (N\times C)}$ is the bias term.
For the optimization, we minimize the differences between the real land-use configurations and the generated land-use configurations, the optimization objective is as follows: 
% \vspace{-0.3cm}
\begin{equation}
   \mathcal{L}_S =  \sum_{k=1}^{K} ||\mathbf{\widehat{X}}^{(k)}-\mathbf{\tilde{X}}^{(k)}||^{2}
   \label{ls}
\end{equation}

\begin{figure*}[hbtp!]
% \vspace{-0.2cm}
	\subfigure[AVG\_KL]{\includegraphics[width=4.35cm]{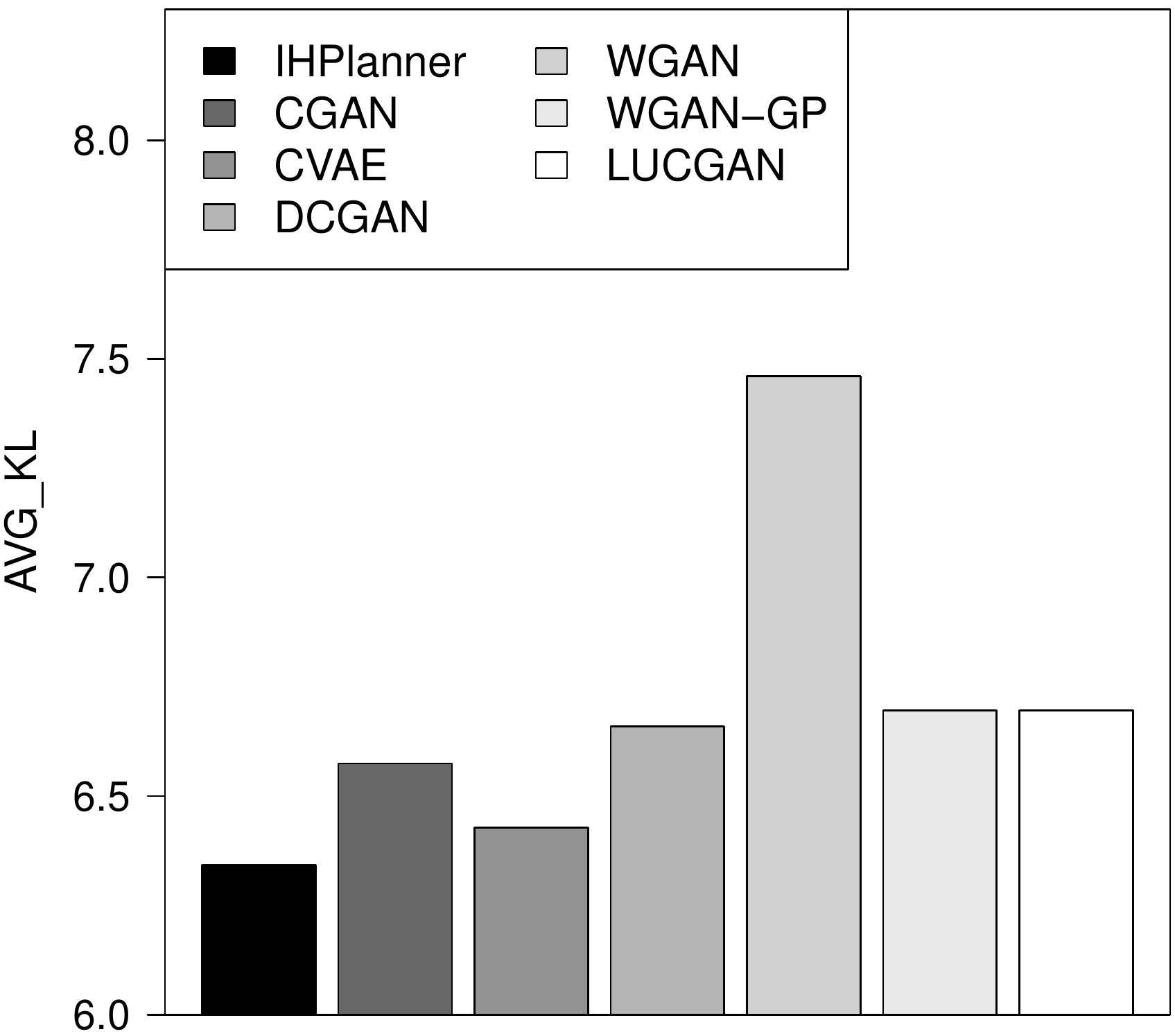} \label{fig:over_avg_kl}}
	\subfigure[AVG\_JS]{\includegraphics[width=4.35cm]{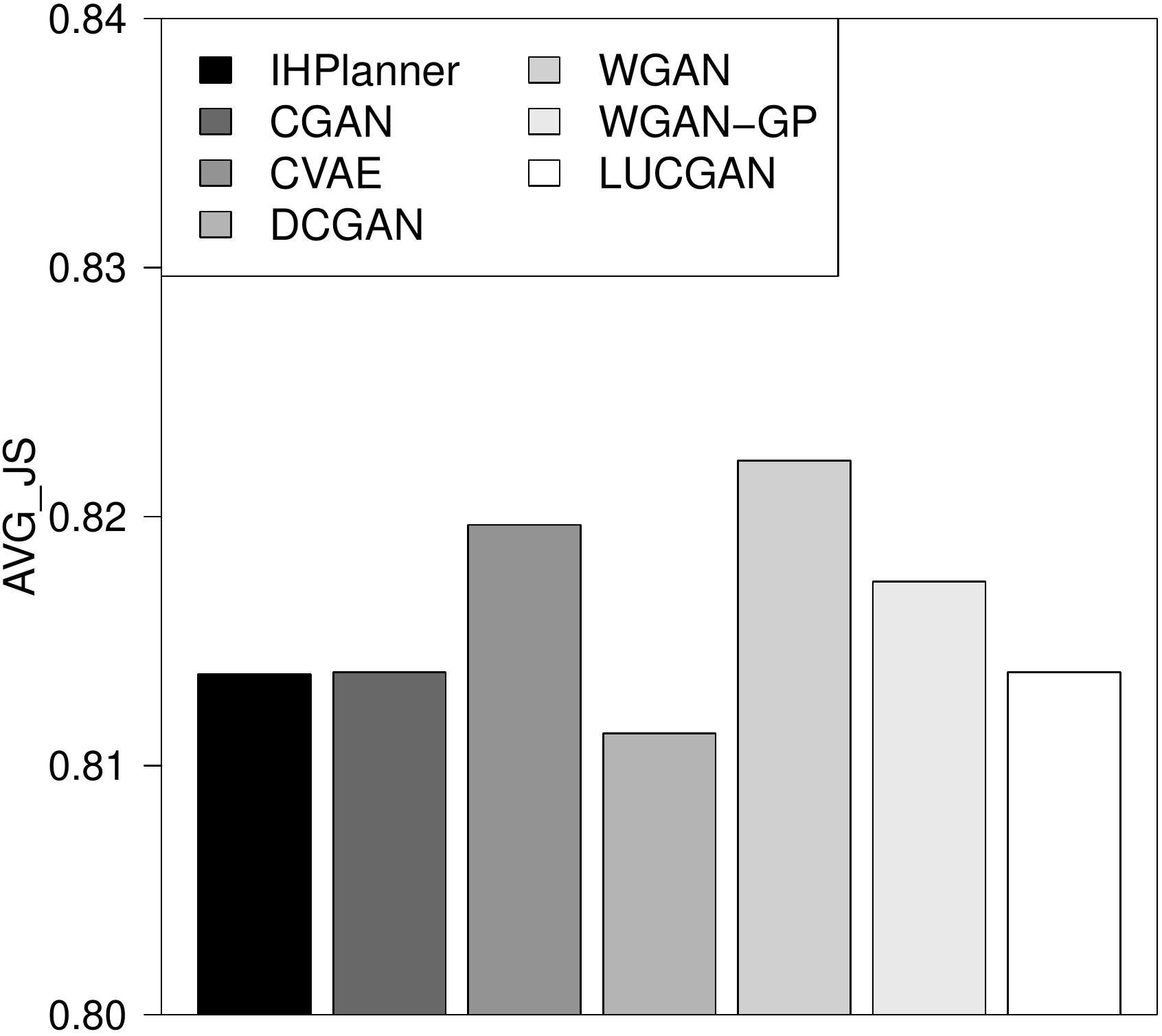}\label{fig:over_avg_js}}
	\subfigure[AVG\_HD]{\includegraphics[width=4.35cm]{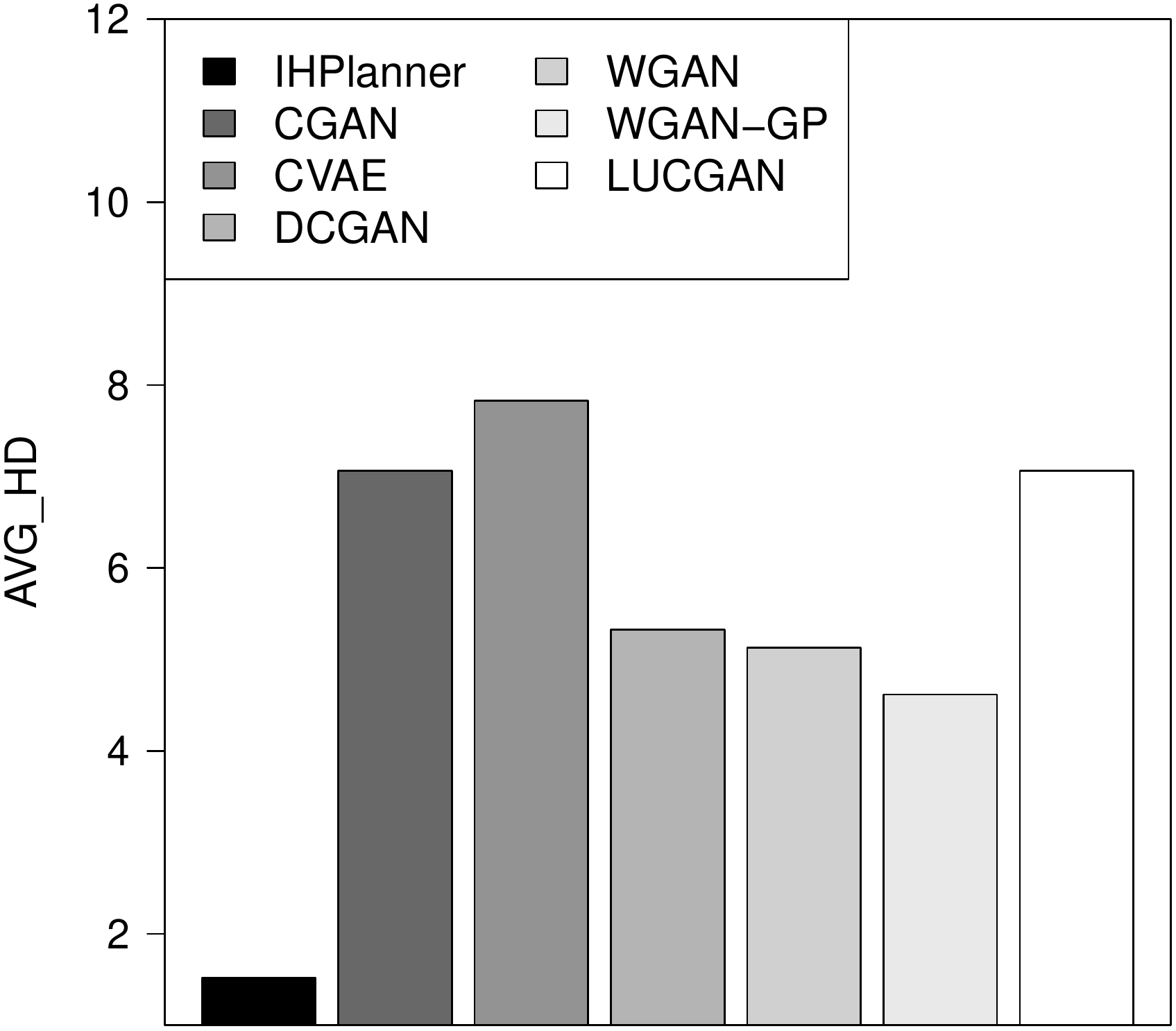}\label{fig:over_avg_hd}}
	\subfigure[AVG\_Cos]{\includegraphics[width=4.35cm]{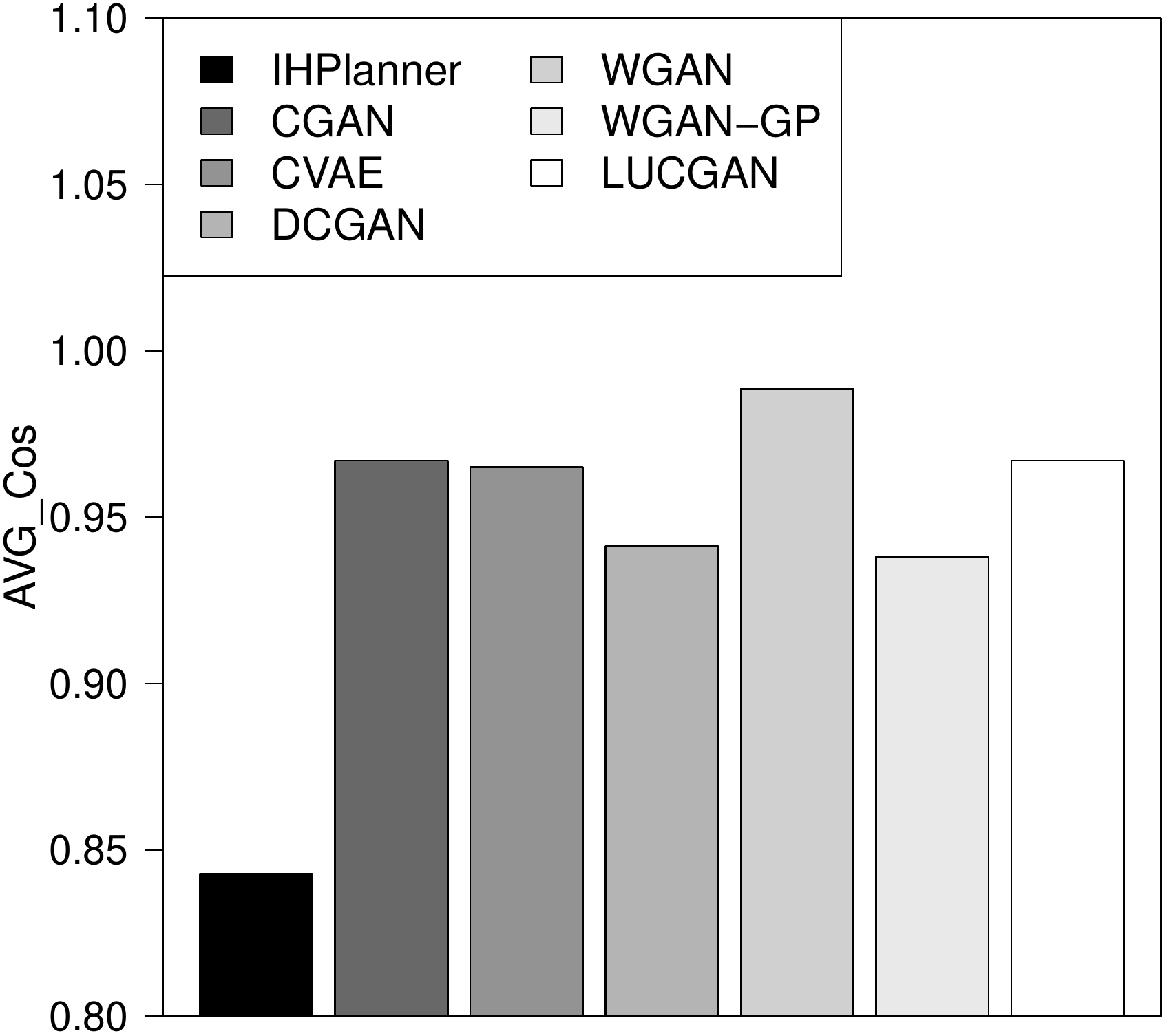}\label{fig:over_avg_Cos}}	
% 	\vspace{-0.25cm}
	\caption{Overall Performance in terms of all evaluation metrics.}
	\label{fig:overall performance}
% 	\vspace{-0.4cm}
\end{figure*}

\begin{figure*}[hbtp!]
% 	\vspace{-0.2cm}
	\subfigure[AVG\_KL]{\includegraphics[width=4.35cm]{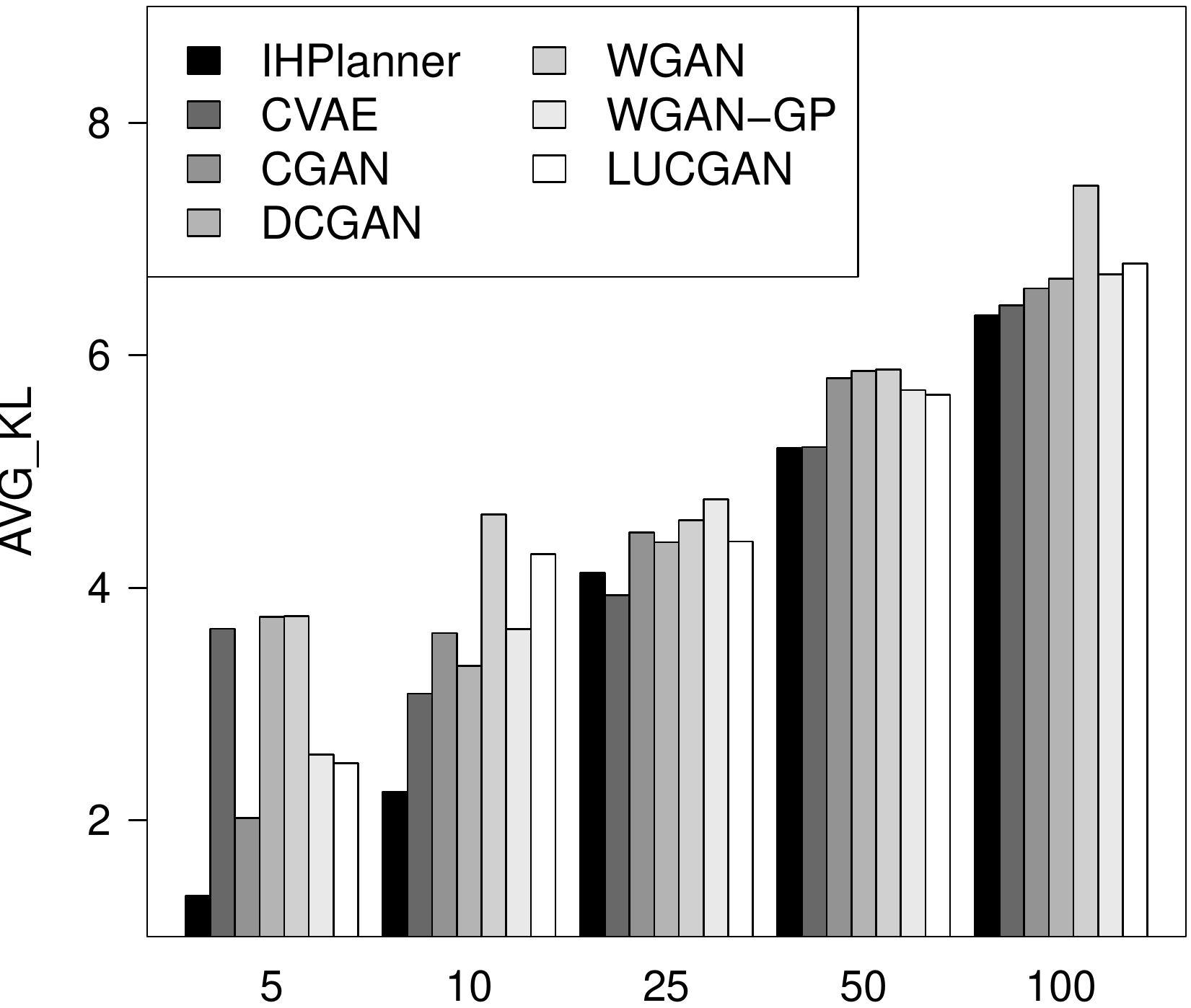} \label{fig:ss_avg_kl}}
	\subfigure[AVG\_JS]{\includegraphics[width=4.35cm]{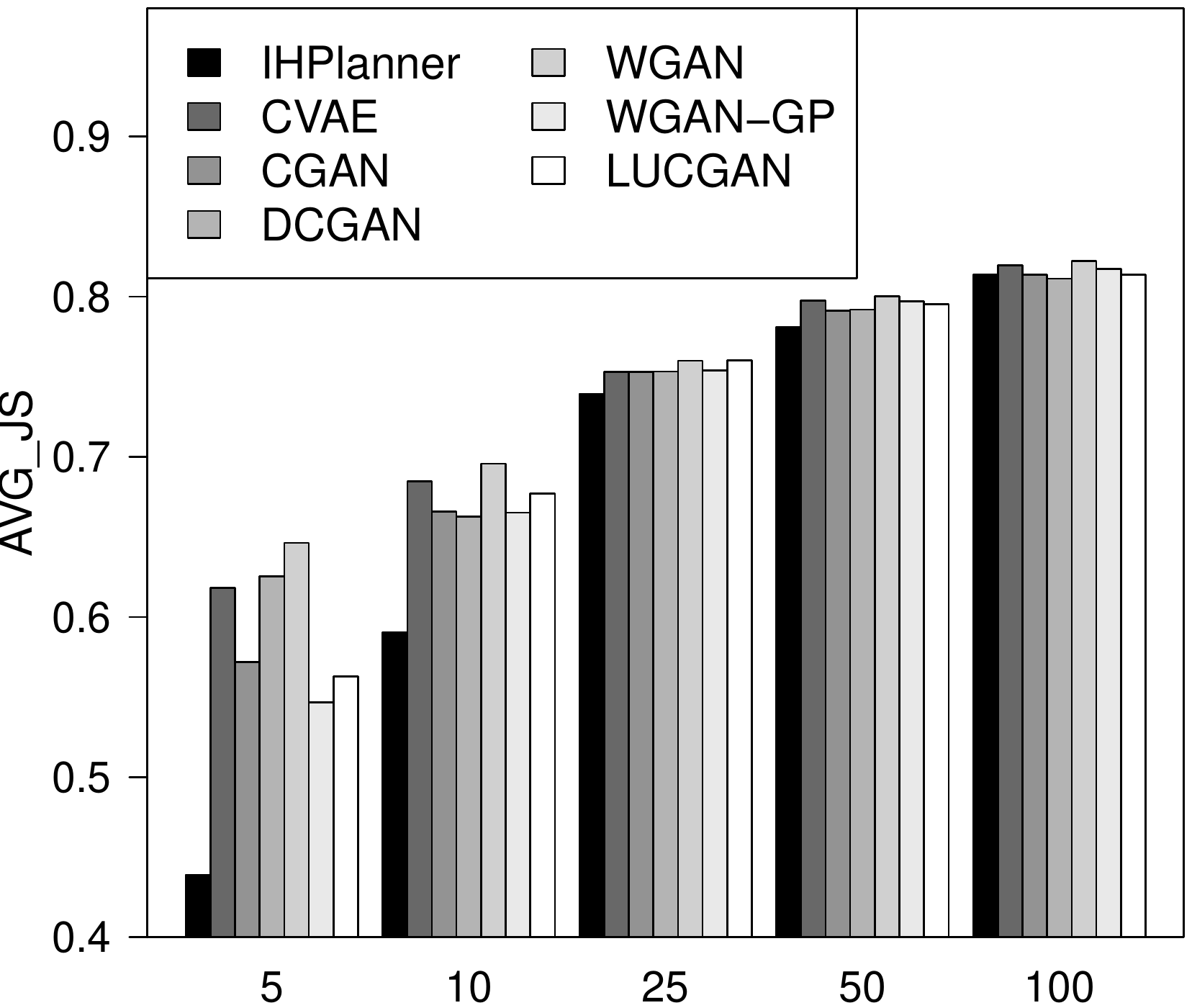}\label{fig:ss_avg_js}}
	\subfigure[AVG\_HD]{\includegraphics[width=4.35cm]{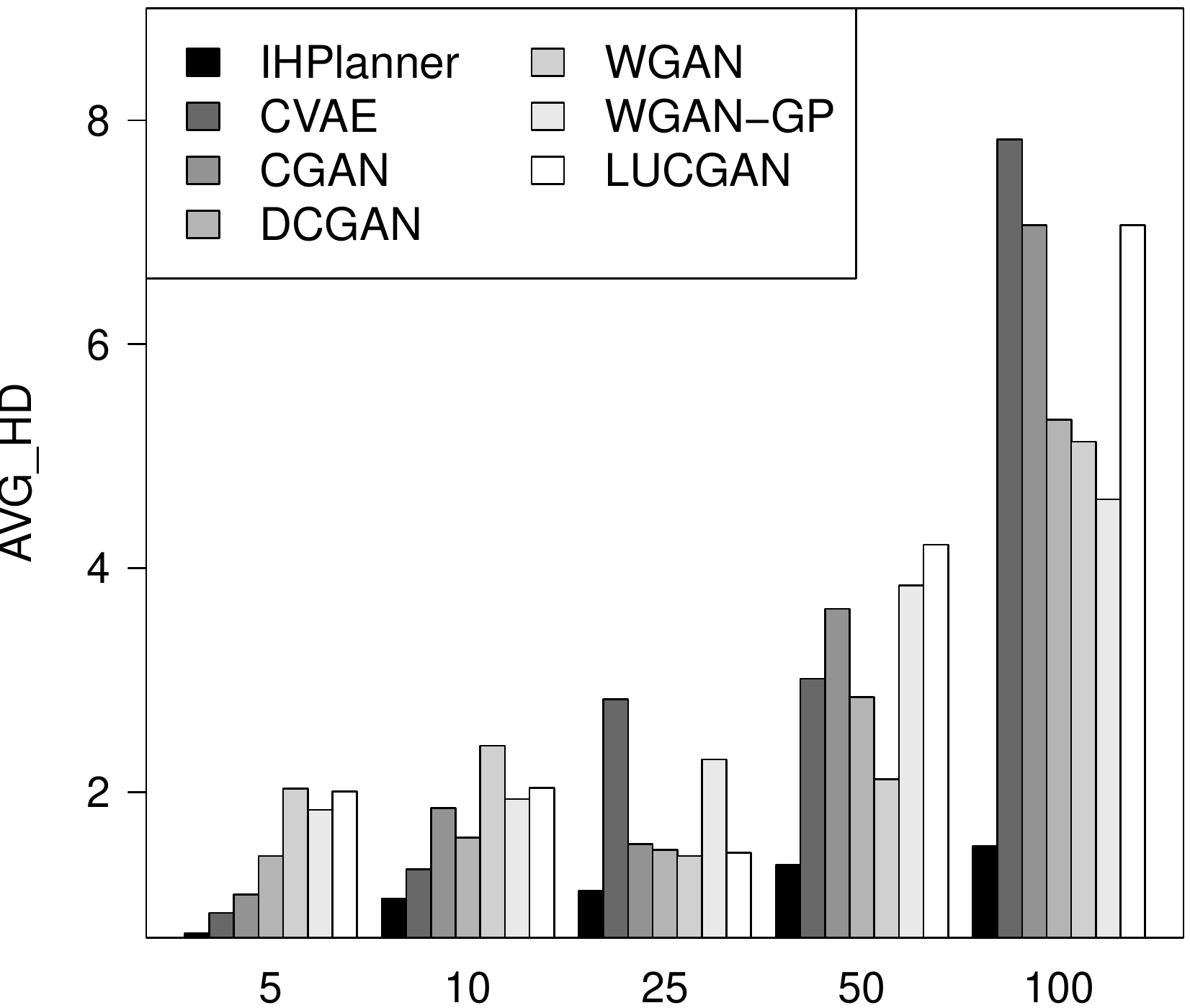}\label{fig:ss_avg_hd}}
	\subfigure[AVG\_Cos]{\includegraphics[width=4.35cm]{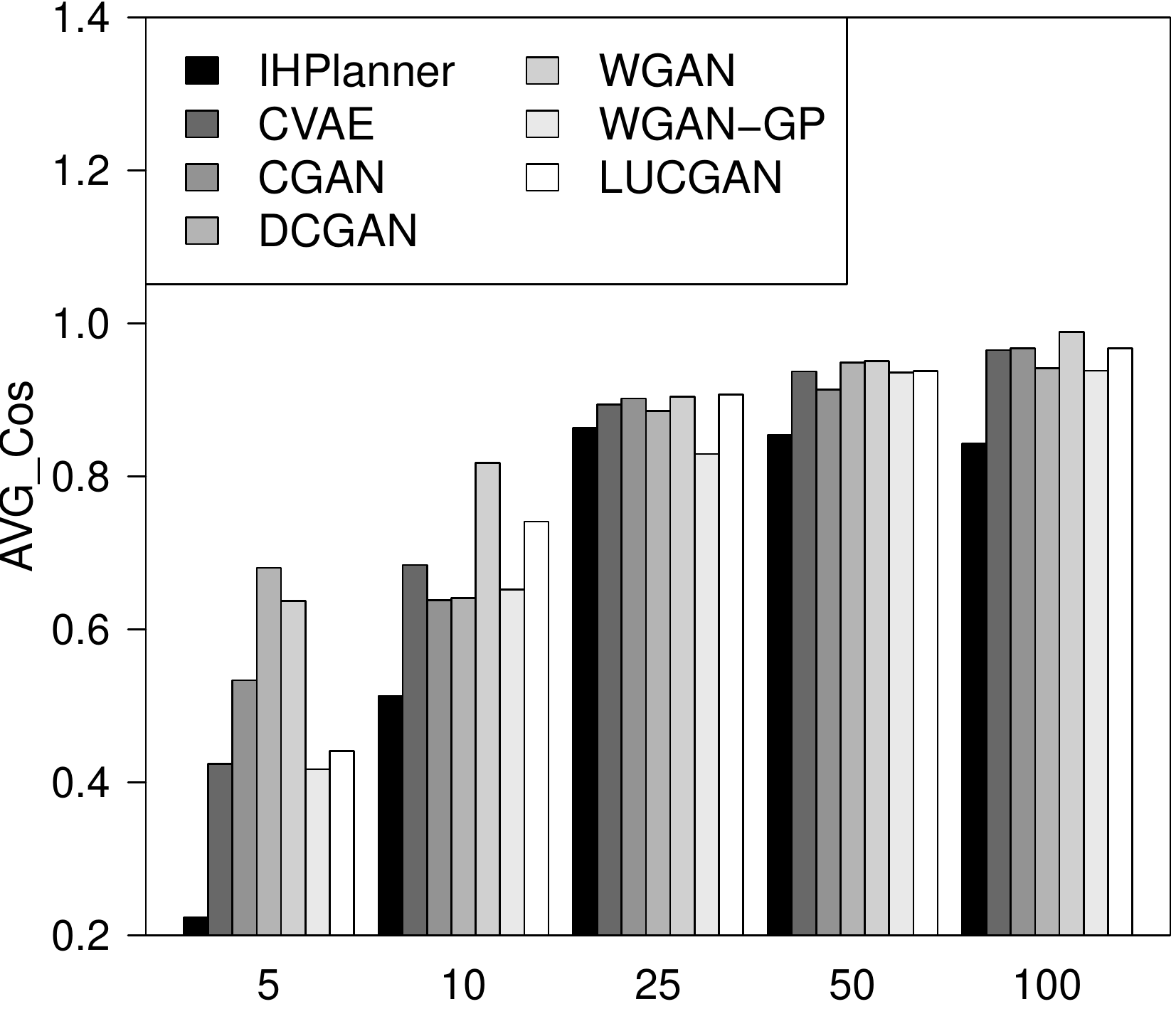}\label{fig:ss_avg_Cos}}	
% 	\vspace{-0.4cm}
	\caption{The influence of different settings of $N$ for land-use configuration generation.}
	\label{fig:square_size}
% 	\vspace{-0.43cm}
\end{figure*}

% \vspace{-0.1cm}
\section{Experiments}
\subsection{Experimental Setup}
\subsubsection{Data Description.}
% Our dataset is composed of land-use configuration, urban functional zones, geospatial contexts, and human instructions.
Our research focuses on Beijing.
The data collection process is as follows:
we first crawled 2990 residential communities from soufun.com and downloaded 328,668 POIs from openstreetmap.org to construct land-use configuration samples referring to ~\cite{wang2020reimagining}.
The categories of these POIs are listed in Table \ref{poi_lists}.
Then, we collected taxi trajectories from T-drive project ~\cite{yuan2010t-drive} and downloaded road networks, POIs from openstreetmap.org. 
to discover urban functional zones  referring to ~\cite{yuan2014discovering}.
Next, we used housing price data crawled from soufun.com, mobile checkins crawled from weibo.com, taxi trajectories, and POIs to extract socioeconomic features of geospatial contexts.
Moreover, we utilized the green rate including in crawled residential community data to construct human instructions.
% Finally, according to the location of each residential community, the land-use configurations, urban functional zones, geospatial contexts, and human instruction were paired to form the final dataset.
% We have provided the download link of the processed dataset and corresponding code in the abstract section.

\begin{table}[htbp!]
\scriptsize
% \vspace{-0.1cm}
\centering
\setlength{\abovecaptionskip}{0.cm}
\caption{POI categories}
\setlength{\tabcolsep}{1mm}{
\begin{tabular}{cccccc}  
\toprule
 code  & POI category & code & POI category & code & POI category \\  
\midrule       
  0  & road & 1 & car service & 2 & car repair \\
  3 & motorbike service & 4 & food service & 5 & shopping \\
  6 & daily life service & 7 & recreation service & 8 & medical service \\
  9 & lodging & 10 & tourist attraction & 11 & real estate \\
  12 & government place & 13 & education & 14 & transportation \\
  15 & finance & 16 & company & 17 & road furniture \\
  18 & specific address & 19 & public service  & & \\
\bottomrule
\end{tabular}}
\label{poi_lists}
% \vspace{-0.2cm}
\end{table}

\subsubsection{Evaluation Metrics}
% In our experimental settings, human instruction refers to the green rate level that reflects the coverage of green plants in a geographical area.
% There are five green rate levels in our dataset:
There are five human instructions (i.e., green rate level) in our dataset: 
 Green0, Green1, Green2, Green3, Green4. 
From left to right, the green rate of the land-use configuration increases.
To assess the generation performance quantitatively, we adopted distribution distances as the evaluation metrics.
The reason is that the data distribution of land-use configurations can be divided into different parts according to human instructions.
Our planner generates a land-use configuration based on a specific human instruction.
Thus, the generated configuration should be close to its green rate level's data distribution part and far from other parts.
Motivated by this idea, we used four evaluation metrics:
1) \textbf{Average Kullback-Leibler (KL) Divergence ~\cite{kullback1951information}:}
$\text{AVG\_KL} = \frac{\sum_{j=1}^{5}w_j\cdot KL(P_j,\widehat{P}_j) }{\sum_{j=1}^{5}w_j}. $ 
2) \textbf{Average Jensen-Shannon (JS) Divergence ~\cite{1207388}:} 
$\text{AVG\_JS} =  \frac{\sum_{j=1}^{5}w_j\cdot JS(P_j,\widehat{P}_j) }{\sum_{j=1}^{5}w_j}.$
3) \textbf{Average Hellinger Distance (HD) ~\cite{Hellinger+1909+210+271}:} $\text{AVG\_HD} =  \frac{\sum_{j=1}^{5}w_j\cdot HD(P_j,\widehat{P}_j) }{\sum_{j=1}^{5}w_j}.$
4)  \textbf{ Average Cosine Distance (Cos) ~\cite{singhal2001modern}:} 
$ \text{AVG\_Cos} =  \frac{\sum_{j=1}^{5}w_j\cdot Cos(P_j,\widehat{P}_j) }{\sum_{j=1}^{5}w_j}.$
In all metric equations, $j$ denotes the human instruction, $w_j$ is the number of land-use configurations belonging to $j$; 
$P_j$ denotes the distribution of original configurations of $j$;
$\widehat{P}_j$ indicates the distribution of generated configurations of $j$;
For all four metrics, the lower the metric value is, the better the generation performance is.

\subsubsection{Baseline Models}
\label{baseline_models}
IHPlanner was compared with the following baseline models:
\textbf{LUCGAN:}~\cite{wang2020reimagining} can generate  an urban plan for a geographical area according to the socioeconomic features of geospatial contexts.
\textbf{CGAN:}~\cite{mirza2014conditional} can create ideal data samples (\textit{e.g.} image, text, speech) based on conditional inputs.
\textbf{CVAE:}~\cite{sohn2015learning} is similar to CGAN yet replacing the generative model with variational autoencoder.
\textbf{DCGAN:}~\cite{radford2015unsupervised} is a classical image generation framework, and it has been adopted into spatiotemporal domain to capture geospatial patterns.
\textbf{WGAN:}~\cite{pmlr-v70-arjovsky17a} is an enhanced GAN, which overcomes the instability of the classical GAN and accelerates it.  
\textbf{WGAN-GP:}~\cite{gulrajani2017improved} 
is an enhanced WGAN, which uses gradient penalty to replace  weights clipping for improving stability.
Besides, we developed four model variants to conduct ablation studies: i)
\textbf{IHPlanner$^{-}$} removes the conditioning augmentation module;
ii) \textbf{IHPlanner$^{*}$} removes the multi-head attention module;
iii) \textbf{IHPlanner$^{'}$} removes the input of human instruction;
iv) \textbf{IHPlanner$^{+}$} removes the input of geospatial contexts.
We randomly split the dataset into two independent sets.
The prior 90$\%$ is the train set, and the remaining 10$\%$ is the test set.
We provided other experimental details in the technical appendix.

% We provided experimental details (\textit{e.g.} hyperparameter settings, experimental platform, etc) in the technical appendix to improve the reproducibility of IHPlanner.

% \vspace{-0.25cm}

\begin{figure*}[htbp]
\centering
% \vspace{-0.3cm}
\subfigure[Original land-use configuration under different human instructions]{
\begin{minipage}[t]{1.0\linewidth}
\centering
\includegraphics[width=7in]{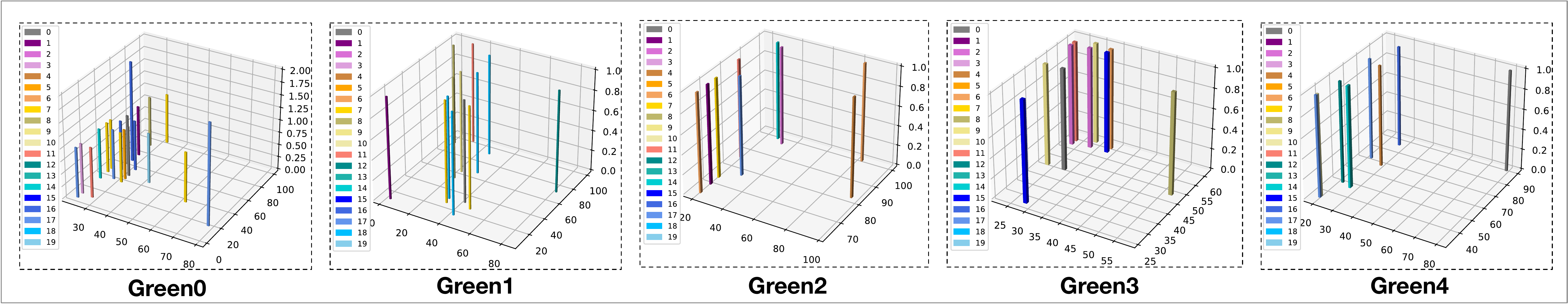}
%\caption{fig1}
\label{fig:original_land}
\end{minipage}%
}%
% \vspace{-0.25cm}

\subfigure[Generated land-use configuration under different human instructions]{
\begin{minipage}[t]{1.0\linewidth}
\centering
\includegraphics[width=7in]{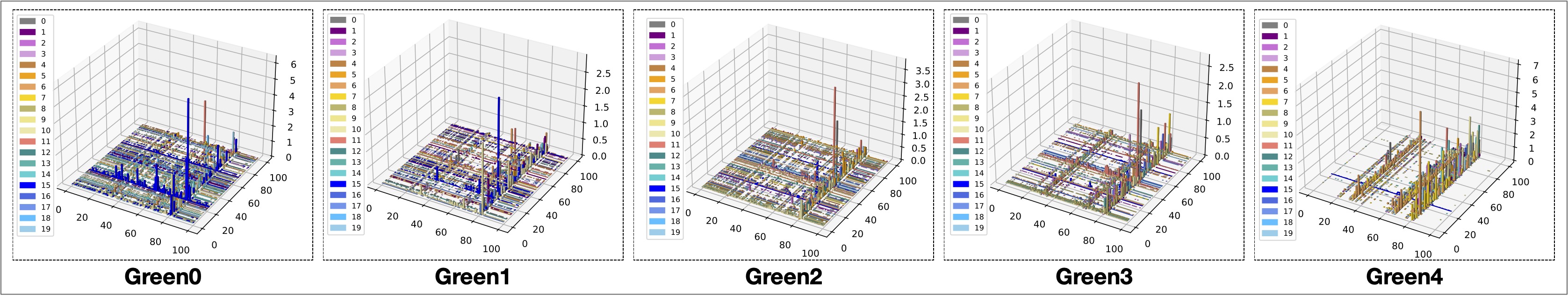}
\label{fig:generate_land}
% \caption{fig2}
\end{minipage}
}%
\centering
% \vspace{-0.35cm}
\caption{Visualization comparison between original land-use configurations and generated land-use configurations }
\label{fig:visual_land}
% \vspace{-0.55cm}
\end{figure*}

% \vspace{-0.3cm}
\subsection{Experimental Results}

% \vspace{-0.3cm}
\subsubsection{Overall Comparison}
This experiment aims to answer: \textit{Can our method (IHPlanner) effectively generate land-use configurations considering human instructions and geospatial contexts?}
Figure \ref{fig:overall performance} shows the overall comparison results in terms of all evaluation metrics.
We observed that IHPlanner outperforms all baseline models.
There are two underlying drivers: i) the functionalizer module effectively projects the planning semantics of human instructions and geospatial contexts into  urban functionality projections. 
Such projections help IHPlanner to understand the planning semantics further and generate human-friendly and environment-friendly urban plans.
ii) By taking into account hierarchical zone-grid and hierarchical zone-zone dependencies in planning, IHPlanner gains suitable generation constraints for developing desirable land-use configurations.

\subsubsection{Robustness Check}
This experiment aims to answer: \textit{Is IHPlanner robust and stable when confronted with different-scale land-use configuration generation tasks?}
We validated the robustness of IHPlanner by changing the value of $N$ that is used to partition the geographical area from $5$, to $10$, to $25$, to $50$, to $100$, respectively.
The greater the value of $N$ is,  the  finer the land-use configuration is.
Figure \ref{fig:square_size} shows the comparison results in terms of all evaluation metrics.
We noticed that IHPlanner outperforms all baseline models regardless of  the value of $N$.
This observation indicates that IHPlanner is more effective in perceiving the requirements of urban planning, resulting from the urban projection process of the functionalizer.
Thus, our method can keep the excellent and robust generation performance.
% Moreover, we observed that as the value of $N$ increases, the generation performance of IHPlanner downgrades relatively.
% A potential reason is that the fine-grained land-use generation task requires capturing more planning details, which raises planning difficulties and thus results in poorer performance.

\subsubsection{Visualization Analysis of the Generated Land-use Configurations.}
% This experiment aims to answer: \textit{Can land-use configurations generated by IHPlanner provide interpretations for real urban planning?}
Figure  \ref{fig:visual_land} illustrates the visualizations of original and generated land-use configurations.
In each subfigure, the left color legend provides the mapping correlation between POI categories and  colors;
the right 3D space exhibits the POI distribution of a land-use configuration sample;
the height of each color bar shows the POIs number at the corresponding location;
the text label under each subfigure is the associated human instruction.
We found that the generated configurations are more organized and capture more planning details than original ones.
In the meantime, we observed that as the green rate level increases, business-related POIs (e.g., POI category 15, 16) decrease, and tourism-related POIs (e.g., POI category 7, 10) increase.
Thus, this observation indicates that IHPlanner can perceive human instructions and surrounding contexts to produce excellent urban plans providing insights to urban experts.

\section{Related Works}
% \vspace{-0.1cm}
\textbf{Deep Generative Learning.}
There are three kinds of popular approaches in the deep generative learning domain: normalizing flows (NF), variational autoencoders (VAE), and generative adversarial networks (GAN). 
NF refers to a set of generative models 
with tractable distributions where both sampling and density evaluation can be efficient and exact ~\cite{kobyzev2020normalizing}.
VAE is capable of learning the latent representations of data and providing deep inference models ~\cite{kipf2016variational}.
GAN is able to simulate the distribution of real data by the competing of  generator and discriminator under a zero-sum game setting ~\cite{creswell2018generative}. 
% Our IHPlanner applies deep generative techniques to the automated urban planning domain, which first generates coarse-grained zones, then produces fine-grained solutions.

% Deep generative
% learning has been successfully applied to many applications ~\cite{zhang2017stackgan,kang2018conditional}.
% yeh2017semantic,qiu2017deep
% For instance, Zhang {\it{et al.}} utilized a stack GAN structure to generate realistic images according to text descriptions ~\cite{zhang2017stackgan}.
% Kang {\it{et al.}} built a semi-supervised VAE structure to design new molecules with desired properties ~\cite{kang2018conditional}. 
% chenthamarakshan {\it{et al.}}  proposed the generation framework CogMol to design the new drug molecules targeting Covid19 ~\cite{chenthamarakshan2020target}.
% Compared with these existing works, our automated urban planner IHPlanner is a pipeline deep generative framework.
% We divide the generation process into two stages: coarse-grained generation, which generates the rough sketch of urban planning;
% fine-grained generation, which produces the detailed urban planning solutions.

\noindent\textbf{Attention Models.}
Attention mechanism gradually becomes a necessary technical module in novel deep neural networks for improving model performance ~\cite{wu2020visual}.
% Attention-based deep models have been successfully applied to many research  domains such as natural language processing, computer vision, graph systems, etc
% % velivckovic2017graph,vaswani2017attention,velivckovic2017graph
% ~\cite{10.1007/978-3-030-67070-2_3,chefer2021transformer,Lee_2018_ECCV}.
For instance, 
% Zhao {\it{et al.}} proposed a novel pixel attention scheme that produces 3D attention maps for generating better super-resolution images ~\cite{10.1007/978-3-030-67070-2_3}. 
~\cite{Lee_2018_ECCV} presented a stacked cross attention framework to discover the latent alignments between the image space and the text space for conducting more accurately image-text matching.
Wang  {{\it et al.}} provided a knowledge graph (KG) attention network that captures the high-order connectivity of KG to improve the recommendation performance ~\cite{wang2019kgat}.
% In this paper, to capture the planning dependencies among different urban subareas, we employ the multi-head attention mechanism.
% Extensive experimental results validate the necessity the attention  component of IHPlanner.

\noindent\textbf{Urban Planning.}
With the popularity of the concept of smart city, urban planning plays a more important role in the urban development ~\cite{wang2018learning,wang2021measuring}. 
% garcia2020sensitivity,,
For instance, ~\cite{khansari2014impacting} studied the impact of the smart city on urban sustainability and urban planning.
% ~\cite{thite2011smart} studied the implications of urban planning for human resource development.
% However, traditional urban planning is a complicated and time-consuming process.
Recently, the remarkable success of deep learning has led researchers to think about how to utilize artificial intelligence to improve the efficiency of urban planning   ~\cite{shen2020machine}.
For example,
% For example, ~\cite{albert2018modeling} exploited a GAN model to learn the urban designing patterns based on real urban footprint images.
~\cite{shen2020machine} utilized a GAN model to fill the urban elements in road map figures to produce the final urban plan.
Compared with these works, IHPlanner is more advanced automatically and practically.

% \vspace{-0.1cm}

% \vspace{-0.15cm}
\section{Conclusion Remarks}
% \vspace{-0.1cm}
In this paper, we propose a revolutionary deep urban planner, namely IHPlanner.
To develop practical planning solutions based on planning requirements, we automate the urban planning process using a hierarchical generation method inspired by the workflow of urban experts.
The input of IHPlanner is the integrated embedding of human instructions and geospatial contexts, which makes IHPlanner able to produce desirable urban plans according to human intention.
The Functionalizer is a significant innovation in IHPlanner, which perceives the planning dependencies between different urban zones via the multi-attention mechanism.
Extensive experiments and case studies demonstrate the effectiveness and superiority of IHPlanner.
In the future, we plan to add more human-machine interactions to make the automated urban planner become more practical.

\section{Acknowledgments}
This research was partially supported by the National Science Foundation (NSF) via the grant numbers: 2040950, 2006889, 2045567.

\bibliography{aaai23}

\end{document}